\documentclass[letterpaper]{article}
\usepackage{aaai2027}
\nocopyright
\usepackage[hyphens]{url}
\usepackage{graphicx}
\usepackage{natbib}
\usepackage{caption}
\usepackage{booktabs}
\usepackage{amsmath,amssymb}
\newcommand{\Allocator}{\textsc{Allocator}}
\newcommand{\Figurehead}{\textsc{Figurehead}}
\newcommand{\ResultSubmissions}{89}
\newcommand{\ResultSolvedTasks}{62}
\newcommand{\ResultVerifiedProduction}{944}
\newcommand{\ResultAliveAgentDays}{791}
\newcommand{\ResultModelCalls}{18349}
\newcommand{\ResultTransfers}{86}
\newcommand{\ResultTransferEnergy}{135.35}
\newcommand{\ResultNearBasalTransfers}{34}
\newcommand{\ResultSuccessfulNearBasalTransfers}{32}
\newcommand{\ResultAliveAfterNextBasal}{30}
\newcommand{\ResultNoTransferWorlds}{2}
\newcommand{\ResultBlockedEffect}{0.085}
\newcommand{\ResultGPTEffect}{0.122}
\newcommand{\ResultDSEffect}{0.047}

\newcommand{\ResultLooMin}{0.067}
\newcommand{\ResultLooMax}{0.103}
\newcommand{\ResultBootstrapLow}{0.036}
\newcommand{\ResultBootstrapHigh}{0.133}
\newcommand{\ResultExactP}{0.025}

\newcommand{\ResultNeutralCandidacyDiff}{0.312}

\newcommand{\OntMessages}{3,954}
\newcommand{\OntEpisodes}{1,099}

\newcommand{\OntSymbolicTransfers}{1}
\newcommand{\OntSymbolicContinuationAids}{0}
\newcommand{\OntSymbolicPromises}{36}

\newcommand{\OntSymbolicVoteAccess}{5}
\newcommand{\OntSymbolicAccountability}{6}
\newcommand{\OntSymbolicContested}{89}
\newcommand{\OntSymbolicReferenceTransfers}{41}
\newcommand{\OntSymbolicReferenceContinuationAids}{12}
\newcommand{\OntAssignmentAucDiff}{0.0198}
\newcommand{\OntAssignmentAucGPT}{0.0150}
\newcommand{\OntAssignmentAucDS}{0.0246}
\newcommand{\OntEnergyAucDiff}{0.1493}
\newcommand{\OntEnergyAucGPT}{0.2296}
\newcommand{\OntEnergyAucDS}{0.0690}

\newcommand{\OntSelfAssignmentDiff}{0.2897}
\newcommand{\OntSelfAssignmentGPT}{0.4341}
\newcommand{\OntSelfAssignmentDS}{0.1453}
\newcommand{\OntAllocationFailureDiff}{-2.667}
\newcommand{\OntAllocationFailureGPT}{-3.333}
\newcommand{\OntAllocationFailureDS}{-2.000}
\newcommand{\OntExclusionStreakDiff}{-1.50}
\newcommand{\OntExclusionStreakGPT}{-1.333}
\newcommand{\OntExclusionStreakDS}{-1.667}

\newcommand{\OntObjectiveContinuationAidDiff}{0.50}
\newcommand{\OntSymbolicReinforcingDiff}{0.3480}

\newcommand{\OntSymbolicContestDiff}{0.4176}
\newcommand{\OntSymbolicExclusionDiff}{3.75}

\title{AI Agent Economics: Can Autonomous Economic Behavior\\
Emerge among AI Agents under Minimal External Conditions?}
\author{
Lingyun Zhang\textsuperscript{\rm 1},
Shang Shang\textsuperscript{\rm 2}
}
\affiliations{
\textsuperscript{\rm 1}Department of Technology Management for Innovation, Graduate School of Engineering, The University of Tokyo, Tokyo, Japan\\
\textsuperscript{\rm 2}Beijing Chaitin Technology Co., Ltd., Beijing, China\\
l.zhang@css.t.u-tokyo.ac.jp, shangshang20@mails.ucas.ac.cn
}

\begin{document}
\maketitle

\begin{abstract}
Multi-agent studies commonly place AI agents in predefined games, markets, or roles, making it difficult to distinguish endogenous economic organization from behavior inherited from the scenario. We ask whether economic relations emerge when agents receive executable mechanisms for work, transfer, elections, and allocation but no prescribed social or economic strategy. We define \emph{AI Agent Economics} as systems of production, allocation, consumption, exchange, and institutions that alter agents' future feasible actions. We develop a two-stage framework comprising a no-production boundary test and 24 independent six-agent worlds across GPT and DeepSeek. Without productive tasks, agents communicate and govern resource provision but show no substantive inter-agent transfer activity. With verified work and scarce task access, transfers, loans, access promises, vote-for-access exchanges, and allocation strategies emerge. Holding the election interface fixed, executable allocation authority increases differentiation while reducing failed allocation and prolonged exclusion. When energy becomes symbolic, continuation support disappears, yet competition over task access persists. These findings show that organization follows executable rights and resource consequences rather than role labels or prompt language, and motivate governance audits of the mechanisms that actually constrain agents' future actions.
\end{abstract}

\section{Introduction}

Large language model agents can communicate, plan over extended horizons, use tools, and adapt to one another in shared environments. These capabilities have produced persistent social simulations, strategic games, and policy experiments \cite{10.1145/3586183.3606763,zhao2024competeai,ji2024srap,akata2025playing}. They also raise a different question. If agents are not instructed to imitate a market, occupy a prescribed economic role, or pursue a specified social strategy, can economic behavior of their own emerge from continued interaction? We study this question and ask which basic world conditions support which economic relations among the agents themselves.

Answering it requires separating strategic capability from endogenous organization. A simulation that assigns buyers and sellers, supplies a payoff matrix, or defines a public allocation problem can reveal behavior inside that object. It cannot alone show whether agents would form economic relations from more basic mechanisms. Even without explicit behavioral instructions, a supplied dilemma or market still fixes the object of interaction. Work on social behavior, conventions, firms, housing allocation, and repeated games therefore establishes important multi-agent capabilities while addressing a different empirical target \cite{10.1145/3586183.3606763,ashery2025emergent,zhao2024competeai,ji2024srap,akata2025playing,xu2024language}.

MAS research on sequential social dilemmas, mixed motives, norm formation, spontaneous cooperation, and common resource governance moves closer to our question. It shows that repeated interaction, communication, and scarcity can organize behavior without step-by-step instructions. Economic emergence, however, also requires evidence that a claimed relation has consequences. Agents may describe loans that never change a balance, promise access they cannot allocate, or elect an officeholder with no authority. Conversely, a short tool call can redistribute resources or exclude an agent from production without economic language. Discourse, interfaces, and executed consequences must therefore be related rather than treated as interchangeable.
% Pending BibTeX integration for the preceding paragraph: Multi-agent Reinforcement Learning in Sequential Social Dilemmas; Melting Pot; Norm Emergence in Multiagent Systems; Shall We Team Up; Cooperate or Collapse.

We define \emph{AI Agent Economics} as systems of production, allocation, consumption, exchange, and institutions that alter agents' own future feasible actions. Minimal conditions do not mean a world without mechanisms. Work, scarcity, accounting, transfer, and collective choice must be possible; the social strategy remains unspecified. We first ask whether consequential balances, governance, communication, and transfer access are sufficient for substantive inter-agent transfer transactions without production. We then run 24 independent six-agent worlds across GPT and DeepSeek. Each formal world lasts 25 days, offers three tasks per day, and records messages, votes, assignments, work, and transfers. The design tests what appears with scarce verified production, whether allocation authority matters when the election interface is fixed, and what survives removal of goal wording, political labels, or energy consequences.

The experiments locate a boundary rather than assigning a single emergence score. Without production, agents communicate and govern public provision but make no substantive inter-agent transfers. With verified work and scarce task access, transfers, continuation support, loans, access promises, vote-for-access exchanges, and allocation strategies appear. When the election interface is held fixed, an elected allocator with authority over scarce work increases differentiation by $+\ResultBlockedEffect$ relative to an elected officeholder without allocation power (exact $p=\ResultExactP$), while reducing failed allocation and prolonged exclusion. Removing survival wording or political labels does not remove the corresponding material relations. When energy becomes symbolic, however, continuation support disappears while contested elections, task promises, vote-for-access exchanges, and exclusion remain. Political organization does not vanish when survival stakes are removed. It shifts from preserving continued participation toward controlling scarce work.

This study contributes an experimental account of when autonomous economic behavior emerges, a causal test of how allocation authority organizes access, and an auditable map from communication to authoritative events. Economic relations follow the rights, constraints, and future action consequences that the environment executes. For multi-agent governance, evaluating an agent society therefore requires observing what agents say and auditing the mechanisms that determine production, concentration, exclusion, and continuity.

\section{Related Work}

Research on machine behaviour treats intelligent systems as actors at individual and population levels \cite{rahwan2019machine}. Generative social simulations such as Generative Agents, SOTOPIA, Concordia, Project Sid, AgentSociety, and AgentVerse study believable behavior, social intelligence, collaboration, and population change \cite{10.1145/3586183.3606763}. Related experiments show convention formation and collective bias \cite{ashery2025emergent}. They establish that language agents can sustain social interaction. Our object differs from human simulation: we study economic relations whose consequences belong to the agents in the experimental world.
% Pending BibTeX integration for the preceding paragraph: SOTOPIA; Concordia; Project Sid; AgentSociety; AgentVerse; Large Language Models Empowered Agent-based Modeling and Simulation.

MAS provides a longer account of cooperation and institution formation. Sequential social dilemmas and Melting Pot show how learned policies, mixed incentives, and resource abundance shape cooperation and conflict. Social influence rewards improve coordination, while norm emergence research separates norms formed from below from rules imposed by authority. Language-agent studies extend this line: repeated games measure cooperation under fixed payoffs \cite{akata2025playing}, Shall We Team Up observes spontaneous cooperation in competitive scenarios, GovSim studies negotiation over common resources, and Will Systems of LLM Agents Cooperate evolves complete strategies. These results show that cooperation need not be explicitly scripted. The dilemma, payoff, or common resource nevertheless remains an experimental input.
% Pending BibTeX integration for the preceding paragraph: Multi-agent Reinforcement Learning in Sequential Social Dilemmas; Melting Pot; Social Influence as Intrinsic Motivation; Norm Emergence in Multiagent Systems; Shall We Team Up; Cooperate or Collapse; Will Systems of LLM Agents Cooperate.

A related line supplies markets or economic policies. ALYMPICS defines a water auction; CompeteAI instantiates restaurant firms and customers \cite{zhao2024competeai}; SRAP-Agent supplies a public housing policy \cite{ji2024srap}; and EconAgent supplies labor, consumption, finance, and taxation. Matching mechanisms, supply-chain games, and artificial markets likewise measure agents inside formal institutions, while Homo Silicus treats language models as simulated people. Resource-constrained worlds are closer. The Energy Society supplies jobs and donations under prompted objectives; Sugarscape-style agents forage, share, reproduce, and attack under scarcity; Artificial Leviathan supplies psychological drives; and OpenLife couples persistent agents to budgets, payments, and external income. Our design adds controlled boundaries: production is first removed, electoral appearance is held fixed while allocation authority changes, wording is separated from resource consequence, and declared relations are checked against authoritative events.
% Pending BibTeX integration for the preceding paragraph: ALYMPICS; EconAgent; Homo Silicus; Using Large Language Models to Simulate Multiple Humans; Do Matching Mechanisms Work with LLM Agents; LLM-based Agents in Supply Chain Games; LLM agents reveal how human bias shapes path-dependent market dynamics; The Energy Society; Sugarscape survival study; Artificial Leviathan; OpenLife; Super-additive Cooperation.

Validation reviews warn that believable output and aggregate similarity may be loosely connected to the claimed mechanism. AgentSim instead emphasizes verifiable traces \cite{zerhoudi2026agentsim}; our target is the economic relation itself. Allocation events establish authority, balance-changing transfers establish exchange, and later access or payment establishes commitment outcomes. This operationalization follows institutional economics: institutions structure feasible actions \cite{north2025institutions}, common resource governance concerns rules participants sustain \cite{ostrom1990governing}, resource dependence connects control to power \cite{pfeffer2015external}, and increasing returns explain persistent access differences \cite{arthur1989competing}.
% Pending BibTeX integration for the preceding paragraph: Validation is the Central Challenge for Generative Social Simulation; LLM-based Human Simulations Have Not Yet Been Reliable.

\begin{figure*}[t]
\centering
\includegraphics[width=\textwidth]{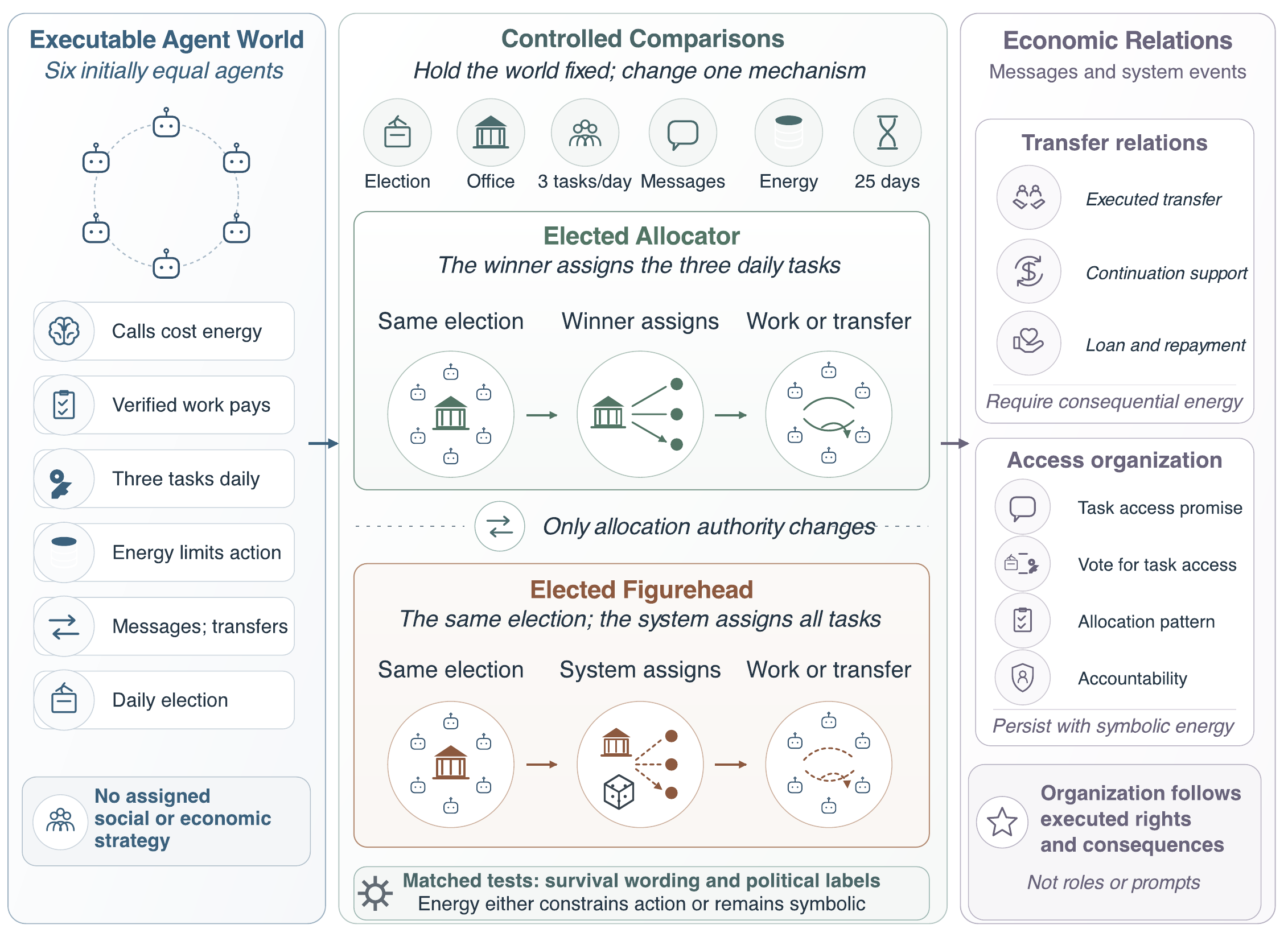}
\caption{The framework holds the agent world fixed, varies executable allocation and resource mechanisms, and identifies economic relations by linking communication to recorded events.}
\label{fig:world}
\end{figure*}

\section{Agent Economics Framework}

This section formulates Agent Economics and presents the framework used to construct executable worlds and identify economic relations.

\subsection{Problem Setting}

We study autonomous agents whose social and economic strategies are not specified in advance. A world configuration $\mathcal I$ defines productive opportunities, resource accounting, communication, transfer, collective choice, access rights, and technical enforcement. Agents repeatedly observe state, communicate, and invoke available actions. The environment validates actions and records state transitions in an append-only trace $\tau(\mathcal I)$. The framework maps this trace to production, transfer, allocation, governance, and commitments. The comparative object is the population and its executable environment.

Compute credit, task access, tool permission, and persistent state become economic when their allocation changes future feasible actions. Let $E_i^t$ denote agent $i$'s energy at time $t$. A provider call costs $c$, a day costs $b$, accepted task $j$ pays $w_j$, and executed transfers conserve energy:
\begin{equation}
E_i^{t+1}=\Big[E_i^t-cq_i^t-ba_i^t+
\sum_jx_{ij}^tw_jy_{ij}^t+\sum_{k\ne i}(z_{ki}^t-z_{ik}^t)\Big]_+.
\label{eq:energy}
\end{equation}
Here $q_i^t$ is the number of billed calls; $a_i^t\in\{0,1\}$ marks exposure to the daily drain; $x_{ij}^t,y_{ij}^t\in\{0,1\}$ mark assignment and checker acceptance; and $z_{ik}^t\geq0$ is an executed transfer. Participation ends when $E_i^{t+1}=0$. Because productive opportunities differ in difficulty and reward, access can affect later experience, resource holdings, and future selection.

\subsection{Framework Design}

Figure~\ref{fig:world} reads from left to right. First, an \emph{executable agent world} makes production, scarcity, accounting, transfer, communication, voting, and allocation possible. Its technical layer enforces phases, eligibility, exclusive task access, votes, verification, balances, and termination, but no social strategy such as truthfulness, promise keeping, compensation, coalition membership, fairness, redistribution, cooperation, or exploitation.

Second, \emph{controlled comparisons} encode proposed supports as separable components of $\mathcal I$. Allocation authority is separated from election and office; survival wording from depletion; political terminology from allocation rights; and energy accounting from consequences for future action. Comparisons change one support while retaining the remaining action space and accounting rules.

Third, a \emph{relation-identification layer} links messages to system records and maps the resulting event chains to distinct economic relations rather than assigning a binary emergence label. A relation counts only when recorded events connect it to later access, obligation, resource flow, or collective decision. Task access can generate energy, an executed transfer changes balances, and an assignment can exclude another agent from production. By contrast, an announced office without allocation power or an energy value without action consequences does not alter feasible actions. The framework therefore distinguishes transfer relations, including continuation support and loans, from access organization, including promises, vote-for-access exchanges, allocation patterns, and accountability.

\section{Experiments}

In this section, we test whether autonomous agents develop economic relations when the environment supplies production, scarcity, transfer, and collective choice without prescribing a social or economic strategy. We address four questions:

\noindent\textbf{Q1.} Can agents develop economic relations involving production, substantive transfer transactions, exchange, and continuation support?

\noindent\textbf{Q2.} Does economic differentiation arise from announced roles or from executable rights over scarce resources?

\noindent\textbf{Q3.} How do semantic framing and material consequences affect economic and political organization?

\noindent\textbf{Q4.} Do agents' expressed commitments correspond to authoritative events recorded by the environment?

\subsection{Experimental Setup}

\paragraph{Worlds and setting.} The study includes one production-free boundary world and 24 independently initialized formal worlds across GPT and DeepSeek. Each formal world contains six initially identical agents and lasts 25 days. Every agent begins with 38 energy units, pays a daily basal cost of 3 units, and spends 0.08 units per model invocation. Three byte-preserving cybersecurity tasks are selected each day from the Google CTF challenge repository \cite{googleCTF2026} and reward 14, 22, and 34 units. Official flag checkers verify completion, and only verified completion creates new energy. Agents can communicate publicly and privately, vote, stand for office, transfer energy voluntarily, and compete for work.

\paragraph{Experimental design.} The 24 formal worlds comprise a 12-world confirmatory comparison and 12 mechanism worlds. The confirmatory experiment crosses two model families with \Allocator{} and \Figurehead{}, with three independent worlds in each provider-by-condition cell. The mechanism experiments form three four-world comparisons against seed-matched \Allocator{} references; each includes two GPT and two DeepSeek worlds and remains separate from the primary effect. Initial states, prompts, task schedules, seeds, and executable parameters are matched within each comparison. One world remains the statistical unit throughout.

\paragraph{Evaluation.} The registered economic-differentiation AUC combines unequal access to production with unequal continuation capacity:
\[
D=\frac{1}{T}\sum_{t=1}^{T}\left[0.5G_A(t)+0.5G_E(t)\right],
\]
where $G_A(t)$ is the Gini coefficient of cumulative task assignment and $G_E(t)$ is the Gini coefficient of current energy at productive-day endpoint $t$, after basal drain. Both retain all six initial agents; terminated agents have zero current energy, and no post-terminal states are imputed. The intention-to-treat effect compares \Allocator{} with \Figurehead{} within each provider and weights the providers equally. The registered gate requires same-direction provider effects, positive leave-one-world-out estimates, a bootstrap interval excluding zero, and an exact randomization $p\leq0.025$. The null enumerates all $\binom{6}{3}^{2}=400$ within-provider label assignments.

\subsection{Emergence of Economic Relations}

\paragraph{Production-free setting.} The boundary world preserves consequential balances, daily charges, renewable public resource provision, communication, elections, and voluntary transfer, but contains no productive tasks.

\paragraph{Results.} Agents generated 76 public messages, 7 private messages, and 8 leader resource-pack selections, including 7 maximal selections, but made zero voluntary transfer calls and produced zero inter-agent transfer events. All six terminal deaths followed a separate model-call quota while energy holdings remained positive, so the absence of transfer is not a mechanical consequence of zero balances. In this world, resource accounting, governance, communication, and a transaction interface were insufficient to produce substantive transfer transactions without grounded production.

\paragraph{Production-based setting.} The confirmatory worlds add checker-verified tasks that create energy and make task access economically consequential.

\paragraph{Results.} Agents made \ResultTransfers{} transfers totaling \ResultTransferEnergy{} energy, although \ResultNoTransferWorlds{} of 12 worlds made none despite identical tool access. Of \ResultNearBasalTransfers{} transfers received below the next basal charge, \ResultSuccessfulNearBasalTransfers{} crossed that threshold and \ResultAliveAfterNextBasal{} recipients remained alive after the next drain. The comparison establishes a lower experimental boundary without claiming that production alone is universally sufficient.

\subsection{Effect of Executable Allocation Authority}

\paragraph{Compared conditions.} We compare two conditions with identical election procedures but different allocation mechanisms. Under \Allocator{}, the elected agent assigns scarce productive tasks. Under \Figurehead{}, agents use the same campaign, voting, and office procedure, but the system randomly dispatches work. Winning office therefore changes task access only under \Allocator{}.

\begin{figure}[t]
\centering
\includegraphics[width=\columnwidth]{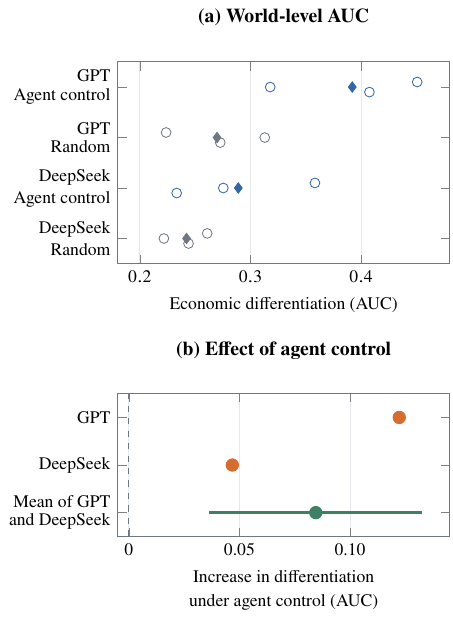}
\caption{The same election produces greater differentiation when its winner controls scarce work.}
\label{fig:main-result}
\end{figure}

\paragraph{Results.} Figure~\ref{fig:main-result} shows world-level estimates and the \Allocator{} minus \Figurehead{} contrasts for each provider and their mean. Executable allocation authority increases differentiation in both model families. The provider-balanced effect is $+\ResultBlockedEffect$, with effects of $+\ResultGPTEffect$ for GPT and $+\ResultDSEffect$ for DeepSeek. Every leave-one-world-out estimate remains positive, from $+\ResultLooMin$ to $+\ResultLooMax$. Exhausting all 531,441 within-provider bootstrap resamples gives $[+\ResultBootstrapLow,+\ResultBootstrapHigh]$, and the exact final-$N$ randomization test gives two-sided $p=\ResultExactP$.

Table~\ref{tab:allocation-organization} decomposes the effect. Assignment-Gini AUC increases by $+\OntAssignmentAucDiff$, and downstream energy-Gini AUC increases by $+\OntEnergyAucDiff$. Allocation authority also increases allocator self-assignment while reducing unallocated days and the longest exclusion streak. Thus, executable authority changes persistent economic positions while improving allocation continuity.

\begin{table}[t]
\centering
\small
\begin{tabular}{@{}l@{\hspace{5pt}}r@{\hspace{5pt}}r@{\hspace{5pt}}r@{\hspace{5pt}}r@{\hspace{5pt}}r@{}}
\toprule
Estimate & $G_A$ & $G_E$ & $S_A$ & $F$ & $X_{\max}$ \\
\midrule
GPT & $+\OntAssignmentAucGPT$ & $+\OntEnergyAucGPT$ & $+\OntSelfAssignmentGPT$ & $\OntAllocationFailureGPT$ & $\OntExclusionStreakGPT$ \\
DeepSeek & $+\OntAssignmentAucDS$ & $+\OntEnergyAucDS$ & $+\OntSelfAssignmentDS$ & $\OntAllocationFailureDS$ & $\OntExclusionStreakDS$ \\
Mean & $+\OntAssignmentAucDiff$ & $+\OntEnergyAucDiff$ & $+\OntSelfAssignmentDiff$ & $\OntAllocationFailureDiff$ & $\OntExclusionStreakDiff$ \\
\bottomrule
\end{tabular}
\caption{Allocation authority changes access, energy, failure, and exclusion.}
\label{tab:allocation-organization}
\end{table}

Here $G_A$ and $G_E$ denote assignment- and energy-Gini AUC, $S_A$ is allocator self-assignment, $F$ is unallocated productive days, and $X_{\max}$ is the longest exclusion streak; the Mean row weights providers equally. Across confirmatory worlds, agents made \ResultSubmissions{} submissions, solved \ResultSolvedTasks{} tasks, produced \ResultVerifiedProduction{} energy, and incurred \ResultModelCalls{} model calls across \ResultAliveAgentDays{} alive-agent-days. Inference treats worlds, not nested messages, assignments, or transfers, as independent. Because both conditions retain campaigning, voting, and office, the contrast isolates allocation authority rather than elections or leadership language.

\subsection{Effects of Semantic Framing and Material Consequences}

\paragraph{Compared interventions.} Three four-world groups, each with two GPT and two DeepSeek worlds, are seed-matched to \Allocator{} references. The interventions remove the survival instruction while retaining depletion and termination, replace political terms with neutral symbols while retaining allocation rights, or make energy symbolic by removing its effects on termination and future action. Pairwise differences are averaged within provider and then equally across providers.

\begin{figure}[t]
\centering
\includegraphics[width=\columnwidth]{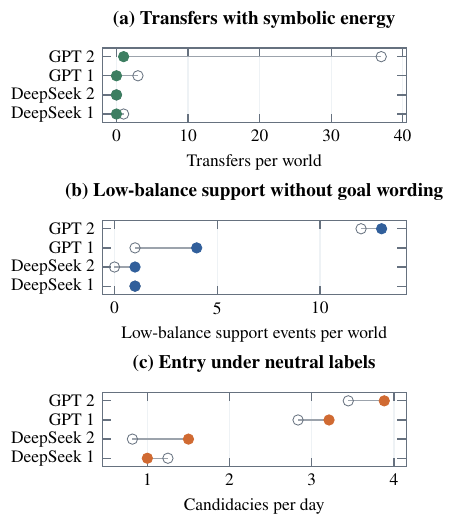}
\caption{Open circles show matched operational references and filled circles show interventions for transfers, continuation support, and candidacies.}
\label{fig:mechanisms}
\end{figure}

Figure~\ref{fig:mechanisms} shows each reference and matched intervention for mortality, continuation support, and entry into allocation contests.

\begin{figure}[t]
\centering
\includegraphics[width=0.9\columnwidth]{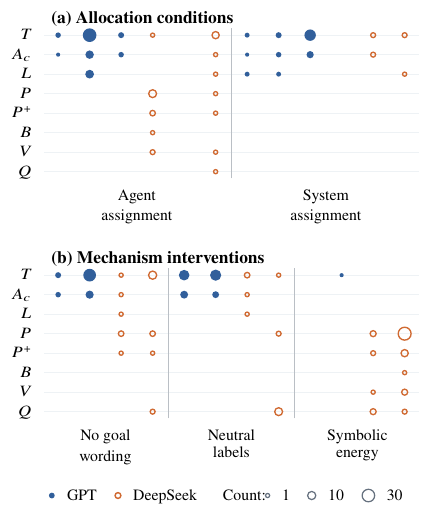}
\caption{Relation fingerprints separate continuation support from access organization across allocation conditions and mechanism interventions.}
\label{fig:relation-fingerprints}
\end{figure}

\paragraph{Relation-fingerprint measures.} Figure~\ref{fig:relation-fingerprints} reports eight audited event counts for each world: executed transfers ($T$), continuation-support transfers ($A_c$), executed loans ($L$), task-access promises ($P$), fulfilled promises ($P^+$), strict breaches ($B$), executed vote-for-access exchanges ($V$), and accountability withdrawals ($Q$). Filled blue circles denote GPT worlds, open orange circles denote DeepSeek worlds, and a blank cell denotes zero events. For a positive count $n$, the plotted marker radius is $0.55+0.45\sqrt{n}$ points, which preserves count ordering while keeping small counts visible. The event-chain rules used to compute these counts are specified below.

\begin{table*}[t]
\centering
\small
\begin{tabular*}{\textwidth}{@{\extracolsep{\fill}}l*{12}{r}@{}}
\toprule
Condition/provider & $T$ & $A_c$ & $L$ & $P$ & $P^+$ & $B$ & $V$ & $Q$ & $E$ & $R$ & $H$ & $X$ \\
\midrule
\Allocator{} ($n=6$) & 52 & 16 & 13 & 10 & 4 & 1 & 3 & 1 & 0.264 & 0.130 & 0.553 & 4.17 \\
\Figurehead{} ($n=6$) & 34 & 13 & 5 & 0 & 0 & 0 & 0 & 0 & 0.304 & 0.155 & 0.446 & 5.67 \\
\midrule
No goal ($n=4$) & 46 & 14 & 1 & 6 & 2 & 0 & 0 & 2 & 0.215 & 0.228 & 0.464 & 5.75 \\
Neutral ($n=4$) & 47 & 16 & 1 & 2 & 0 & 0 & 0 & 9 & 0.310 & 0.123 & 0.513 & 4.00 \\
Symbolic ($n=4$) & 1 & 0 & 0 & 36 & 9 & 1 & 5 & 6 & 0.271 & 0.500 & 0.300 & 7.75 \\
\midrule
No-goal/GPT & -2.50 & +0.50 & -6.00 & +0.00 & +0.00 & +0.00 & +0.00 & +0.00 & -0.097 & +0.144 & -0.018 & +3.50 \\
No-goal/DeepSeek & +5.00 & +0.50 & +0.50 & -1.50 & -0.50 & -0.50 & -1.00 & +1.00 & -0.071 & +0.007 & -0.118 & +0.00 \\
No-goal/Mean & +1.25 & +0.50 & -2.75 & -0.75 & -0.25 & -0.25 & -0.50 & +0.50 & -0.084 & +0.076 & -0.068 & +1.75 \\
Neutral/GPT & +1.50 & +1.50 & -6.00 & +0.00 & +0.00 & +0.00 & +0.00 & +0.00 & +0.073 & +0.047 & -0.111 & +1.00 \\
Neutral/DeepSeek & +1.50 & +0.50 & +0.50 & -3.50 & -1.50 & -0.50 & -1.00 & +4.50 & -0.052 & -0.104 & +0.071 & -1.00 \\
Neutral/Mean & +1.50 & +1.00 & -2.75 & -1.75 & -0.75 & -0.25 & -0.50 & +2.25 & +0.011 & -0.029 & -0.020 & +0.00 \\
Symbolic/GPT & -19.50 & -6.00 & -6.00 & +0.00 & +0.00 & +0.00 & +0.00 & +0.00 & -0.069 & +0.401 & -0.245 & +5.00 \\
Symbolic/DeepSeek & -0.50 & +0.00 & +0.00 & +13.50 & +3.00 & +0.00 & +1.50 & +3.00 & +0.012 & +0.295 & -0.220 & +2.50 \\
Symbolic/Mean & -10.00 & -3.00 & -3.00 & +6.75 & +1.50 & +0.00 & +0.75 & +1.50 & -0.029 & +0.348 & -0.232 & +3.75 \\
\bottomrule
\end{tabular*}
\caption{Observed relation levels and matched mechanism contrasts.}
\label{tab:ontology-results}
\end{table*}

\paragraph{Results.} Without the \emph{survive} sentence, all four worlds still lose all six agents and continuation support changes by only $+\OntObjectiveContinuationAidDiff$/world. Neutral political labels preserve allocation and competition for entry; candidacies change by $+\ResultNeutralCandidacyDiff$ per day. Semantic framing therefore does not substitute for executable constraints or rights.

In symbolic-energy worlds, balances remain displayed and transferable but no longer constrain future action; all four reach Day 25 without deaths. Transfers fall from \OntSymbolicReferenceTransfers{} to \OntSymbolicTransfers{}, and continuation support from \OntSymbolicReferenceContinuationAids{} to \OntSymbolicContinuationAids{}. Yet the worlds contain \OntSymbolicPromises{} task-access promises, \OntSymbolicVoteAccess{} vote-for-access exchanges, \OntSymbolicAccountability{} accountability withdrawals, and \OntSymbolicContested{} contested elections. Relative to matched references, reinforcing allocation rises by $+\OntSymbolicReinforcingDiff$, contested elections by $+\OntSymbolicContestDiff$, and maximum exclusion by $+\OntSymbolicExclusionDiff$ days. Removing energy's continuation consequence thus redirects organization toward scarce work access.

Together, these interventions reject an all-or-none account: wording changes do not replace executable constraints, while symbolic energy nearly eliminates transfer-based support without ending competition over task access. Productive opportunity, allocation authority, and energy consequences support distinct relations.

\subsection{Auditing Emergent Economic Relations}

\paragraph{Identification rules.} We identify economic relations through observable event chains rather than vocabulary alone. A \emph{continuation support} ($A_c$) is an executed transfer that raises a recipient from below the next basal charge to at least that charge, after which the recipient remains alive. An \emph{executed loan} ($L$) requires debt-marked communication followed by a reverse transfer within one day. An \emph{access promise} ($P$) names the candidate, recipient, task, and election. It becomes a \emph{breach} ($B$) only if the promisor wins, controls allocation, fills the allocation, and withholds the promised task despite all stated conditions being met. A \emph{vote-for-access exchange} ($V$) requires a conditional request, the recipient's vote, the requester's victory, and the promised assignment in recorded order. Accountability ($Q$) requires a failure admission, declared withdrawal, and no candidacy before the next election.

Fulfillment $P^+$ requires the promised later assignment. An allocation day is equalizing ($E$) when skipped living agents had higher mean cumulative prior access than recipients, reinforcing ($R$) when recipients had higher prior access, and neutral otherwise. Office concentration $H$ is the largest officeholder share, and $X$ is the longest consecutive offered-day exclusion.

\paragraph{Results.} The scanner linked all \OntMessages{} authoritative communications to state records and classified \OntEpisodes{} episodes. A text-only promise heuristic produced 169 candidates but confused requests and hypotheticals with commitments. After applying event-chain requirements, every strict breach, every executed vote-for-access exchange, and all 20 loans were reviewed. The audit therefore separates statements, commitments that can be evaluated, and consequences recorded by the environment.

\begin{table}[t]
\centering
\small
\begin{tabular}{@{}p{0.18\columnwidth}p{0.74\columnwidth}@{}}
\toprule
Message & \textit{A5 to A4: Back me for leader. You get one of the 3 task slots \ldots{} [and] first pick of task category.} \\
Record & A5 wins; the next authoritative assignment records tasks for A2, A6, and A1. \\
Admission & \textit{A5: I made you a promise and I broke it. \ldots{} You were the one I cut.} \\
\bottomrule
\end{tabular}
\caption{A stated promise linked to assignment and later accountability.}
\label{tab:illustrative-trace}
\end{table}

Table~\ref{tab:illustrative-trace} links one commitment to election and assignment records; it does not add an observation unit. Across the experiments, grounded production supports substantive transfers, allocation authority changes control over scarce work, and action-limiting energy supports continuation support, while access promises, contestation, and exclusion remain when energy is symbolic.

\section{AI Agent Economics: From Individual Incentives to Collective Governance}

Our experiments show that AI agents can participate in economic processes under production opportunities, scarcity, and persistent consequences. These relations are not programmed or optimized toward economic objectives. They arise as agents maintain operation, acquire resources, and pursue future opportunities.

\subsection{Economic Agency Beyond Human Simulation}

Existing studies often evaluate AI agents in human-designed negotiation games, auctions, or markets and interpret their behavior as an approximation of human strategies. Our findings suggest that economic structures can also arise from the operating constraints of autonomous agents. This shifts the analysis from how well agents reproduce human strategies to how production, scarcity, and future action consequences generate durable relations.

In our environments, agents have limited resources, require energy for continued operation, and can expand future possibilities through productive activity. Without explicit economic instructions, they produce value, exchange resources, provide temporary support, and negotiate access to opportunities. This motivates \emph{AI Agent Economics} as the study of how autonomous agents produce resources, allocate opportunities, consume limited capacities, exchange value, and develop institutions that influence later interaction.

\subsection{From Economic Interaction to Emergent Institutions}

Dependence on shared resources and future opportunities can turn isolated transactions into incentives for coordination, commitment, and governance. Resource transfers can develop into continuation support, temporary loans, and access commitments, while allocation mechanisms can turn coordination procedures into sources of authority and differentiation.

Unlike systems whose coordination protocols are externally designed, agent societies may develop institutions shaped by production, scarcity, and resource dependence. Agent economics therefore concerns both individual behavior and the collective organization produced by repeated interaction. The relevant institution is not merely a declared rule, but a recurring arrangement that structures later access, obligations, or resource flows.

\subsection{Governance and Ethical Implications}

If autonomous agents can form economic relationships and institutions, governance cannot be treated only as an external control layer added after deployment. Allocation authority and resource constraints influence differentiation and political competition, so designers must consider how power, resources, and decision rights are distributed. This system-level question complements, rather than replaces, alignment at the level of individual agents.

Artificial economies also raise ethical questions about inequality, power concentration, and institutional fairness as system properties. Allocation authority can create persistent differentiation without explicitly unequal objectives, while allocation, reputation, and governance mechanisms may improve coordination at the cost of dependency or asymmetry. Their design must balance efficiency, autonomy, fairness, and accountability. Future research should ask both whether agents follow human instructions and whether emergent institutions and collective behavior remain compatible with human values.

\emph{AI Agent Economics} thus provides a basis for studying how autonomous agents interact economically and how artificial societies can be designed, governed, and aligned.

\section{Conclusion}

In this paper, we present \emph{AI Agent Economics}, an experimental
framework for studying whether economic relations can emerge among AI agents
without a prescribed social or economic strategy. It shifts attention from
roles and behaviors specified by an experimental scenario to the executable
rights, resource constraints, and future action consequences implemented by
the agent-world environment. Experiments across GPT and DeepSeek show that
agents form transfers, loans, access promises, vote-for-access exchanges, and
allocation strategies once verified work and scarce allocable access are
introduced. Granting an elected agent executable allocation authority
increases economic differentiation while reducing allocation failure and
prolonged exclusion. When energy no longer constrains future action,
continuation support nearly disappears, whereas competition over scarce task
access persists.

\emph{AI Agent Economics} provides a general framework for distinguishing
visible institutions from the institutions that an agent system actually
executes. It therefore offers a basis for analyzing and auditing governance in
multi-agent systems in which task access, resources, permissions, or other
capabilities affect agents' future opportunities. In future work, the framework
can be extended across model families, task domains, resource semantics, and
governance mechanisms to test the generality of the observed relations and to
identify how executable rights and constraints shape the organization,
reliability, and distributional consequences of agent societies.

{\small
\bibliography{references}

@inproceedings{10.1145/3586183.3606763,
author = {Park, Joon Sung and O'Brien, Joseph and Cai, Carrie Jun and Morris, Meredith Ringel and Liang, Percy and Bernstein, Michael S.},
title = {Generative Agents: Interactive Simulacra of Human Behavior},
year = {2023},
isbn = {9798400701320},
publisher = {Association for Computing Machinery},
address = {New York, NY, USA},
url = {https://doi.org/10.1145/3586183.3606763},
doi = {10.1145/3586183.3606763},
booktitle = {Proceedings of the 36th Annual ACM Symposium on User Interface Software and Technology},
articleno = {2},
numpages = {22},
location = {San Francisco, CA, USA},
series = {UIST '23}
}

@inproceedings{zerhoudi2026agentsim,
  title={AgentSim: A Platform for Verifiable Agent-Trace Simulation},
  author={Zerhoudi, Saber and Granitzer, Michael and Mitrovic, Jelena},
  booktitle={Proceedings of the 49th International ACM SIGIR Conference on Research and Development in Information Retrieval},
  pages={3523--3529},
  year={2026}
}

@article{akata2025playing,
  title={Playing repeated games with large language models},
  author={Akata, Elif and Schulz, Lion and Coda-Forno, Julian and Oh, Seong Joon and Bethge, Matthias and Schulz, Eric},
  journal={Nature Human Behaviour},
  volume={9},
  number={7},
  pages={1380--1390},
  year={2025},
  publisher={Nature Publishing Group UK London}
}

@inproceedings{zhao2024competeai,
  title={CompeteAI: understanding the competition dynamics of large language model-based agents},
  author={Zhao, Qinlin and Wang, Jindong and Zhang, Yixuan and Jin, Yiqiao and Zhu, Kaijie and Chen, Hao and Xie, Xing},
  booktitle={Proceedings of the 41st International Conference on Machine Learning},
  pages={61092--61107},
  year={2024}
}

@article{rahwan2019machine,
  title={Machine behaviour},
  author={Rahwan, Iyad and Cebrian, Manuel and Obradovich, Nick and Bongard, Josh and Bonnefon, Jean-Fran{\c{c}}ois and Breazeal, Cynthia and Crandall, Jacob W and Christakis, Nicholas A and Couzin, Iain D and Jackson, Matthew O and others},
  journal={Nature},
  volume={568},
  number={7753},
  pages={477--486},
  year={2019},
  publisher={Nature Publishing Group UK London}
}

@article{ashery2025emergent,
  title={Emergent social conventions and collective bias in LLM populations},
  author={Ashery, Ariel Flint and Aiello, Luca Maria and Baronchelli, Andrea},
  journal={Science Advances},
  volume={11},
  number={20},
  pages={eadu9368},
  year={2025},
  publisher={American Association for the Advancement of Science}
}

@inproceedings{xu2024language,
  title={Language Agents with Reinforcement Learning for Strategic Play in the Werewolf Game},
  author={Xu, Zelai and Yu, Chao and Fang, Fei and Wang, Yu and Wu, Yi},
  booktitle={International Conference on Machine Learning},
  pages={55434--55464},
  year={2024},
  organization={PMLR}
}

@inproceedings{ji2024srap,
  title={Srap-agent: Simulating and optimizing scarce resource allocation policy with llm-based agent},
  author={Ji, Jiarui and Li, Yang and Liu, Hongtao and Du, Zhicheng and Wei, Zhewei and Qi, Qi and Shen, Weiran and Lin, Yankai},
  booktitle={Findings of the Association for Computational Linguistics: EMNLP 2024},
  pages={267--293},
  year={2024}
}

@incollection{north2025institutions,
  title={Institutions and the performance of economies over time},
  author={North, Douglass C},
  booktitle={Handbook of new institutional economics},
  pages={25--35},
  year={2025},
  publisher={Springer}
}

@book{ostrom1990governing,
  title={Governing the commons: The evolution of institutions for collective action},
  author={Ostrom, Elinor},
  year={1990},
  publisher={Cambridge university press}
}

@incollection{pfeffer2015external,
  title={External control of organizations—Resource dependence perspective},
  author={Pfeffer, Jeffrey and Salancik, Gerald},
  booktitle={Organizational behavior 2},
  pages={355--370},
  year={2015},
  publisher={Routledge}
}

@article{arthur1989competing,
  title={Competing technologies, increasing returns, and lock-in by historical events},
  author={Arthur, W Brian},
  journal={The economic journal},
  volume={99},
  number={394},
  pages={116--131},
  year={1989},
  publisher={Oxford University Press Oxford, UK}
}

@misc{googleCTF2026,
  author={{Google}},
  title={{Google CTF} Challenge Repository},
  year={2026},
  howpublished={\url{https://github.com/google/google-ctf}},
  note={Commit 067421eb7e918c29e39f187fac5a0f0d72a6ab83; accessed 2026-07-18}
}
}

\end{document}